\documentclass[sigconf]{acmart}

\AtBeginDocument{%
  }

\newcommand{\eg}{\textit{e.g.,}}
\newcommand{\ie}{\textit{i.e.,}}

\acmConference[ASSETS '24 - The Future of Urban Accessibility Workshop]{Make sure to enter the correct
  conference title from your rights confirmation emai}{October 28--30,
  2024}{Canada, St. John's}

\acmISBN{978-1-4503-XXXX-X/18/06}

\begin{document}

% \title{Research summary of the guide dog robot}
% \title{Guide dog robot development map}
% \title{Guidelines of developing a guide dog robot}
% \title{Lessons learned while developing a guide dog robot}
\title{Lessons Learned from Developing a Human-Centered \\ Guide Dog Robot for Mobility Assistance}

% \author{Hochul Hwang}
% \affiliation{%

\author{Hochul Hwang}
\affiliation{%
  \institution{University of Massachusetts Amherst}
  \city{Amherst}
  \country{USA}}
\email{hochulhwang@cs.umass.edu}

\author{Ken Suzuki}
\affiliation{%
  \institution{University of Massachusetts Amherst}
  \city{Amherst}
  \country{USA}}
\email{kensuzuki@umass.edu}

\author{Nicholas A Giudice}
\affiliation{%
  \institution{University of Maine}
  \city{Orono}
  \country{USA}}
\email{nicholas.giudice@maine.edu}

\author{Joydeep Biswas}
\affiliation{%
  \institution{University of Texas at Austin}
  \city{Austin}
  \country{USA}}
\email{joydeepb@cs.utexas.edu}

\author{Sunghoon Ivan Lee}
\affiliation{%
  \institution{University of Massachusetts Amherst}
  \city{Amherst}
  \country{USA}}
\email{silee@cs.umass.edu}

\author{Donghyun Kim}
\affiliation{%
  \institution{University of Massachusetts Amherst}
  \city{Amherst}
  \country{USA}}
\email{donghyunkim@cs.umass.edu}

% \affiliation{%

\renewcommand{\shortauthors}{Hwang et al.}
\begin{abstract}
 While guide dogs offer essential mobility assistance, their high cost, limited availability, and care requirements make them inaccessible to most blind or low vision (BLV) individuals. Recent advances in quadruped robots provide a scalable solution for mobility assistance, but many current designs fail to meet real-world needs due to a lack of understanding of handler and guide dog interactions. In this paper, we share lessons learned from developing a human-centered guide dog robot, addressing challenges such as optimal hardware design, robust navigation, and informative scene description for user adoption. By conducting semi-structured interviews and human experiments with BLV individuals, guide-dog handlers, and trainers, we identified key design principles to improve safety, trust, and usability in robotic mobility aids. Our findings lay the building blocks for future development of guide dog robots, ultimately enhancing independence and quality of life for BLV individuals.
\end{abstract}



\maketitle

% Experience report 
% 	Aim of this study
% 	What we did so far
%     Remaining challenges

% \begin{figure}[h]
%   \centering
%   \includegraphics[width=.9\linewidth]{figure/gdr-level.pdf}
%   \caption{Development roadmap for the guide dog robot.}
%   \Description{Guide dog robot development level.}
%   \label{f1:roadmap}
% \end{figure}

\section{Introduction}

Over 36 million people globally and one million in the U.S. live with severe visual impairments, a number expected to double within the next 30 years~\cite{ackland2017world,cdc_fast_facts,nih_double}. Guide dogs are one of the most effective mobility aids to support the independent mobility of blind or low vision (BLV) people~\cite{History_Guide_Dogs, hersh2010robotic}. Extensive studies have highlighted the benefits of guide dogs compared to other assistive methods (\eg{} white canes) in terms of mobility, independence, and enhanced confidence~\cite{refson1999health,whitmarsh2005benefits,lloyd2008guide,guide-dog-interview}. However, only a small portion of BLV people have access to guide dogs, mainly due to the limited supply~\cite{howmany-guidedog}. Training a guide dog requires over \$50,000 and approximately two years. The limited supply is further strained by the short working lifespan of guide dogs, typically less than ten years, often resulting in wait times of up to two years for acquisition~\cite{forbes_guidedog}, a challenge that became even more pronounced during the COVID-19 pandemic. Even after adoption, handlers are responsible for lifelong care and maintenance, including medical care, feeding, and daily walks. Additional health and personal factors, such as allergies to animals or limited living space, can further complicate the adoption of a guide dog. Despite their significant benefits, guide dogs may not be a scalable or sustainable solution for the larger population of BLV people.

Motivated by recent progress in quadruped robots and their potential for mass production and long-term sustainability~\cite{unitree_go1, xiaomi_cyberdog, deeprobotics_lite3}, we aim to create a practical guide-dog robot as an additional solution for navigation assistance for BLV people. Unlike wearable or hand-held devices~\cite{smartcane, slade2021multimodal,chuang2018deep, ye2016co, ulrich2001guidecane, takizawa2015kinect, takizawa2012kinect, faria2010electronic, saaid2016smart, belt, headmount, GOnet, zeng2017camera, katzschmann2018safe, li2016isana, rodriguez2012assisting, strumillo2018different, dakopoulos2009wearable}, which require users to scan their surroundings actively, quadruped robots can offer complete local navigation, significantly improving ambulation speed, safety, and comfort~\cite{lloyd2008guide, clark1986efficiency,miner2001experience}. Mobile robots for navigation assistance have been studied for 40 years, beginning with wheeled systems~\cite{Tachi, meldog, tobita2018structure, guerreiro2019cabot, tobita2017examination, kulyukin2004robotic, megalingam2019autonomous, nanavati2018coupled, galatas2011eyedog, kayukawa2019bbeep} and more recently moving to quadruped robots~\cite{xiao2021robotic, chen2022quadruped, hamed2019hierarchical, NSK, due2023walk, cai2024navigating}. Compared to wheeled systems, quadruped robots offer superior traversability that enables navigation through uneven terrains and heterogeneous structures (\eg{} stairs and curbs) in real-world environments. 

Despite a long history of mobile robot research, none have been deployed for BLV individuals due to a limited understanding of how guide dogs and their handlers work together in real life. For example, some robots are too bulky for various situations such as public transportation or crowded areas~\cite{hamed2019hierarchical, NSK, due2023walk} that can be easily seen in urban environments. Also, challenges such as inappropriate leash systems, improper handler positioning, and overemphasis on autonomous navigation without addressing user adoption have hindered development~\cite{xiao2021robotic, hwang2023system}. To understand how navigation is performed by people with BLV, we need to distinguish between the two roles that constitute the navigation task: \emph{Orientation} and \emph{Mobility}~\cite{wiener2010foundations}. Handlers are responsible for \emph{orientation}, which involves knowing and executing routes, and making key navigational decisions, such as turning or crossing a street at waypoints. On the other hand, guide dogs take on the \emph{mobility} role, ensuring the handler's safety by detecting and avoiding collisions with pedestrians or objects such as trash cans, cars, or bicycles~\cite{giudice2010establishing}.

Our prior work investigating the interaction between guide dogs and handlers found that Orientation assistance, similar to what navigation applications (\eg{} Google Maps) provide, is only accepted by people with BLV when there is strong trust, built through long-term positive experiences~\cite{hwang2024towards}. Therefore, achieving autonomy requires overcoming hurdles not only in technology development but also in user adoption. This was a major reason why we first developed a system solely focused on obstacle avoidance without orientation assistance in our prior work~\cite{hwang2023system}. Another important aspect to consider is that the orientation role is gradually delegated to a guide dog as they become familiar with the environment~\cite{hwang2024towards}, which is our next target to be tackled in our project.

\begin{figure}[t]
  \centering
  \includegraphics[width=.9\linewidth]{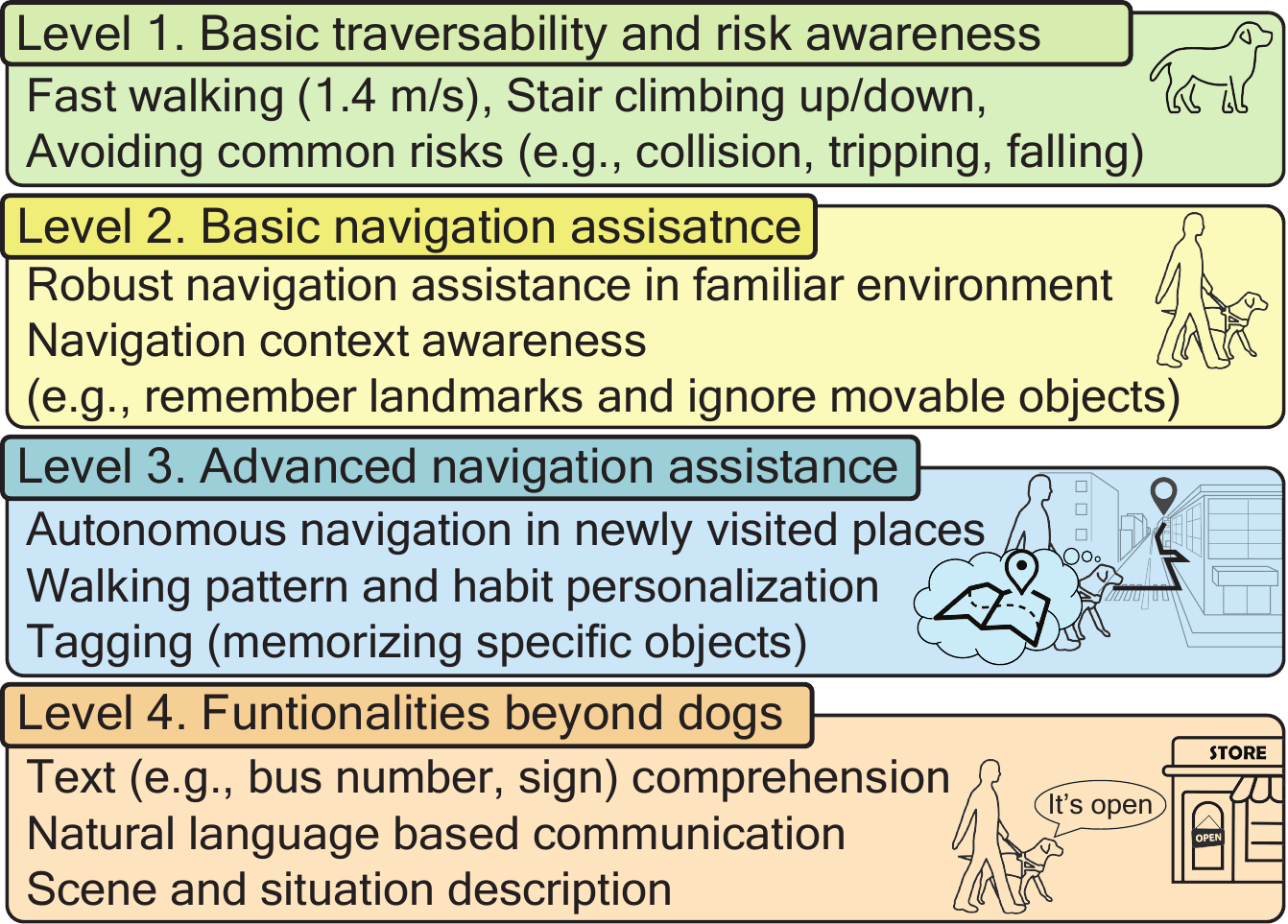}
  \caption{Development roadmap for the guide dog robot.}
  \Description{Guide dog robot development level.}
  \label{f1:roadmap}
\end{figure}

Misunderstandings about basic human-guide dog interaction are also common. For example, systems proposed by \cite{xiao2021robotic, chen2022quadruped, kim2023train} employed a soft leash, which is problematic as handlers must use a rigid harness handle to perceive immediate feedback about the guide dog’s motion~\cite{guerreiro2019cabot}. In another study~\cite{hamed2019hierarchical, cai2024navigating}, the handler was positioned behind the robot during navigation, whereas a guide-dog handler should be positioned next to the dog for safety and efficiency. Some studies address technical problems that are less critical or unimportant for assisting navigation for BLV people. For example, \cite{hamed2019hierarchical} focused on enhancing the locomotion stability of quadruped robots, although our study measuring the pulling forces of guide dogs indicated that even the robot's default controller can sustain the forces~\cite{hwang2023system}.

\begin{figure}[t]
  \centering
  \includegraphics[width=\linewidth]{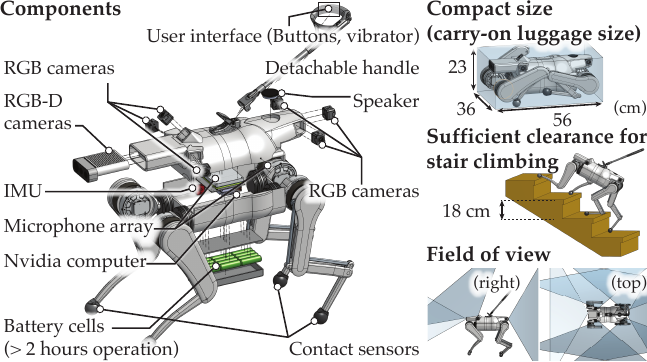}
  \caption{Concept drawing of a guide dog robot.}
  \Description{Guide dog robot design.}
  \label{f1:design}
\end{figure}

To avoid such mistakes and design an effective guide-dog robot, we propose a \emph{human-centered approach} and share insights gained from our previous work, which includes an iterative process of 1) technology development, 2) evaluation via human experiments, and 3) refinement via formative studies. Our prior studies~\cite{hwang2023system, hwang2024towards, hwang2024safe}, including semi-structured interviews with 23 blind guide-dog handlers, 6 BLV individuals, and six experienced trainers, along with several observational and participatory sessions, led us to identify critical design criteria and technical challenges, and build a roadmap for guide-dog robot development as described in Fig.~\ref{f1:roadmap}. With a vision where quadruped robots are ubiquitous, this project ambitiously seeks to establish the standards for guide-dog robot design and development, ultimately enhancing independence and quality of life for BLV people.

\section{User-Centered Insights}
The successful completion of this project aims to shift the focus of guide dog robot research from a technology-centered approach to a human-centered one. Rooted in our comprehensive understanding of the interaction between guide dogs and their handlers, we share lessons learned throughout the development process such as: 1) new quadruped robot hardware featuring compactness, portability, extended operation, and multi-modal sensing; 2) a co-optimization framework for robot hardware and controllers to create an energy-efficient and comfortable navigation assistant for BLV people; and 3) a robust navigation system that adapts to scene variations based on our learning framework utilizing multi-sensory data and foundation models.

\subsection{Optimal guide dog robot hardware and locomotion controller}
Commercially available small-scale quadruped robots~\cite{unitree_go1, xiaomi_cyberdog, deeprobotics_lite3} offer potential as research platforms but face significant limitations as guide dog robots due to their form factor, sensor suite, battery life, and locomotion control. Their short legs limit ground clearance, preventing stair climbing, and larger robots are impractical for public transport, where guide dogs must fit under seats and be portable (\eg{} fit in a carry-on baggage)~\cite{hwang2024towards, guerreiro2019cabot}. Additionally, current sensors lack the field of view needed to detect overhead obstacles or account for occlusions caused by the user's proximity. Battery life is another critical issue—while some robots claim 2–4 hours of operation~\cite{UnitreeGo2}, real-world tests show less than an hour, and added sensors further reduce this~\cite{hwang2023system}. Furthermore, the noise and vibration caused by current locomotion controllers disrupt the auditory and haptic feedback critical for BLV users~\cite{chen2022can}. Addressing these issues, we aim to design a quadruped robot optimized for BLV assistance as depicted in Fig.~\ref{f1:design}.

Recent advancements in reinforcement learning (RL)-based locomotion control~\cite{lee2020learning, miki2022learning, cheng2023parkour, zhuang2023robot, hoeller2023anymal}, have shown impressive agility and robustness in quadruped robots. However, prior work has paid little attention to achieving gentle and quiet walking, and existing RL approaches have not addressed the coupled relationship between hardware design and locomotion control. To optimize both, we believe that a RL training framework that simulates multiple robot designs with randomly sampled parameters, searching for the best locomotion policy in parallel~\cite{rudin2022learning} can iteratively refine design configurations until the optimal hardware and controller for gentle walking are identified.

\subsection{Foundation model-based robust navigation}
Although mobile robot navigation has a rich history of research~\cite{biswas2014vector, biswas2013localization, macenski2023survey, macenski2020marathon2}, existing approaches rely heavily on building detailed maps, continuously updating them, and requiring significant computational resources for real-time localization~\cite{starship}. These requirements pose significant problems, not only because data gathering, labeling, and updating are time-consuming and economically expensive, but also because the high computational demands are difficult to accommodate in a guide-dog robot with limited size and battery power. Additionally, the top-performing algorithms in localization benchmarks such as KITTI~\cite{geiger2012we} rely on LiDAR sensors, which are expensive and heavy for integration into a guide-dog robot. On the other hand, state-of-the-art visual localization and mapping solutions, which do not rely on such sensors, are still not robust enough to handle the scene variations that will be encountered by a guide-dog robot in real-world environments~\cite{vslamComparison2021, rabiee2020ivslam}. Relying on conventional metric localization and mapping for the guide dog robot setting is thus impractical, given the robot's limited computational resources and payload, and the stringent robustness needs of the application.

\begin{figure}[t]
  \centering
  \includegraphics[width=.9\linewidth]{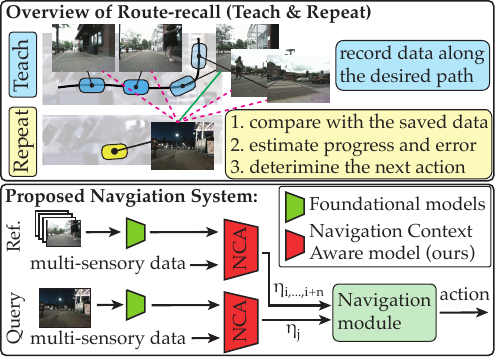}
  \caption{Overview of the proposed navigation system.}
  \label{f3:vtr-pipeline}
\end{figure}

However, we have a key insight that can help us sidestep these challenges. Instead of solving the general-purpose navigation problem (\ie{} driving from any point A to point B), we aim to develop a navigation system capable of recording desired routes and recalling the demonstrated paths without building a meticulously tailored global map or accurately estimating the robot's position. This navigation setup is called the `route-recall' or `visual teach and repeat (VT\&R)' problem~\cite{zhang2009robust} (see Fig.~\ref{f3:vtr-pipeline}). The way VT\&R formulates the navigation task resembles how a guide dog learns a new path: by repeatedly traveling specific routes under the supervision of a sighted person. From an algorithmic perspective, VT\&R relies less on localization accuracy and is computationally cheaper since navigation actions are determined based on memorized scenery and rough position sequences, similar to how humans (or animals) navigate previously visited places~\cite{gothard1996dynamics}. 

While promising, we need to address issues related to scenery variations between demonstration and recall. Reasoning about \emph{semantics} is key to robustness -- when memorizing a path, the robot should focus on objects that are most informative about progress along a route, such as detecting road intersections, notable buildings, or specific stores, while ignoring less informative objects like parked cars or trash cans~\cite{zhu2021lifelong,adkins2022pom}. Recent advances in visual foundational models (\eg{} DinoV2~\cite{oquab2023dinov2}) have shown great promise in understanding general-purpose semantics in images and are robust to lighting variations and viewpoint changes. However, these models are not optimized to understand the \emph{navigation} context, which is crucial for route-recall navigation. To overcome these challenges, we are developing a VT\&R navigation system, using a self-supervised learning framework that integrates foundational models — renowned for their ability to interpret general-purpose semantics and adapt to changes in lighting and perspective — and exploring methods that utilize large-language models (LLMs)~\cite{shah2023lm,anwar2024remembr} and disentangled representation learning that may allow the robot to differentiate between critical and non-essential variations, improving its reliability in dynamic environments and enhancing the practicality of guide dog robots for BLV users.

Unlike fully autonomous guide dog robots, which may not be ideal during initial interactions~\cite{zhang2023follower} due to potential trust biases~\cite{freedy2007measurement}, we believe that a VT\&R pipeline can foster a more natural development of trust. Just as novice guide dog handlers initially struggle to trust their guide dogs but gradually form strong, reliable bonds, we envision a similar trajectory for guide dog robots. Through repeated exposure and consistent performance, this approach has the potential to cultivate trust progressively, ensuring a smoother transition, ultimately enhancing the user experience through autonomy.

% 	- Guidelines for future work	
% 		- System integration and human study

% Indoor local navigation (icra'23)
% - Outdoor navigation has it’s own difficulty

% Decision making based on complex scene understanding

\section{Conclusion}
% In conclusion, this work highlights the crucial need for a human-centered approach to developing guide dog robots, ensuring they meet the real-world needs of blind or low vision (BLV) individuals. By integrating lessons learned from semi-structured interviews, observations of guide dog-handler interactions, and early development phase, we identified key principles for hardware design, navigation pipeline, and user trust that are essential for creating effective robotic mobility aids. The insights shared here provide a foundation for future research, including optimizing quadruped robot form factors and leveraging advanced learning models to improve scene understanding and route recall. Our findings offer a clear roadmap for the continued development of guide dog robots, with the potential to significantly enhance the independence and quality of life for BLV individuals through scalable, reliable mobility solutions.

In conclusion, this work highlights the crucial need for a human-centered approach in developing guide dog robots to meet the real-world needs of blind or low vision (BLV) individuals. By integrating insights from interviews, guide dog-handler observations, and early development, we identified key principles in hardware design, navigation, and user trust. These findings provide a foundation for future research, focusing on optimizing quadruped robots and leveraging advanced learning models for robust navigation. Our findings offer a clear roadmap for the continued development of guide dog robots, with the potential to significantly enhance the independence and quality of life for BLV individuals through scalable and reliable mobility solutions.

%  Introduction
% 	- Why not delivery robot?
% 	- Why quadruped robot?

% Why? Challenges: hardware, perception, 
% - hardware
%     - dimension
%     - battery life
%     - dust proof, (water proof)
%     - wheel legged, legged 
% - perception
%     - intelligent disobedience - ref) seeing eye paper
% - navigation
%     - recongnize scene
%     - control
%     - action model
%     - data 
% - Trust

% - VLMs for navigation and decision making
% 	- VTR
% 	- Scene description
% - Lessons from observations
% 	- GDs are not perfect, we should aim for 100\%, but can be deployed even when it is not 100\% correct all the time
% - CYBATHLON Challenge

% Functionalities
% 	- basic obedience
% 	- scene description

% \subsection{Findings}

% \begin{table}
%   \caption{Urban vs. suburban}
%   \label{tab:gdts}
%   \begin{tabular}{cccccc}
%     \toprule
%     \textbf{} & \textbf{Age} & \textbf{Gender} & \textbf{Hello} & \textbf{Interview} & \textbf{Observation} \\
%     \midrule
%     Challenges & 65 & M & 45 & 2 & 1 \\
%     T02 & 52 & M & 19 & 3 & 5 \\
%     T03 & 58 & M & 35 & 4 & 5 \\
%     \bottomrule
% \end{tabular}
% \Description{Difficulty of t}\\
% \end{table}

% \subsection{Discussion}
% \subsection{Conclusion}

\begin{acks}
The study materials used in this study have been reviewed and approved by the University of Massachusetts Amherst IRB, Federal Wide Assurance \# 00003909 (protocol ID: 3124, 5224, and 5709).
\end{acks}

\bibliographystyle{ACM-Reference-Format}
\bibliography{sample-base}


\begin{thebibliography}{81}

%%% the Web address.

\ifx \showCODEN    \undefined \def \showCODEN     #1{\unskip}     \fi
\ifx \showDOI      \undefined \def \showDOI       #1{#1}\fi
\ifx \showISBNx    \undefined \def \showISBNx     #1{\unskip}     \fi
\ifx \showISBNxiii \undefined \def \showISBNxiii  #1{\unskip}     \fi
\ifx \showISSN     \undefined \def \showISSN      #1{\unskip}     \fi
\ifx \showLCCN     \undefined \def \showLCCN      #1{\unskip}     \fi
\ifx \shownote     \undefined \def \shownote      #1{#1}          \fi
\ifx \showarticletitle \undefined \def \showarticletitle #1{#1}   \fi
\ifx \showURL      \undefined \def \showURL       {\relax}        \fi
% invisible to TeX
\providecommand\bibfield[2]{#2}
\providecommand\bibinfo[2]{#2}
\providecommand\natexlab[1]{#1}
\providecommand\showeprint[2][]{arXiv:#2}

\bibitem[Ackland et~al\mbox{.}(2017)]%
        {ackland2017world}
\bibfield{author}{\bibinfo{person}{Peter Ackland}, \bibinfo{person}{Serge Resnikoff}, {and} \bibinfo{person}{Rupert Bourne}.} \bibinfo{year}{2017}\natexlab{}.
\newblock \showarticletitle{World blindness and visual impairment: despite many successes, the problem is growing}.
\newblock \bibinfo{journal}{\emph{Community eye health}} \bibinfo{volume}{30}, \bibinfo{number}{100} (\bibinfo{year}{2017}), \bibinfo{pages}{71}.
\newblock


\bibitem[Adkins et~al\mbox{.}(2022)]%
        {adkins2022pom}
\bibfield{author}{\bibinfo{person}{Amanda Adkins}, \bibinfo{person}{Taijing Chen}, {and} \bibinfo{person}{Joydeep Biswas}.} \bibinfo{year}{2022}\natexlab{}.
\newblock \showarticletitle{Probabilistic Object Maps for Long-Term Robot Localization}. In \bibinfo{booktitle}{\emph{Intelligent Robots and Systems (IROS), IEEE/RSJ International Conference on}}. IEEE.
\newblock


\bibitem[Anwar et~al\mbox{.}(2024)]%
        {anwar2024remembr}
\bibfield{author}{\bibinfo{person}{Abrar Anwar}, \bibinfo{person}{John Welsh}, \bibinfo{person}{Joydeep Biswas}, \bibinfo{person}{Soha Pouya}, {and} \bibinfo{person}{Yan Chang}.} \bibinfo{year}{2024}\natexlab{}.
\newblock \showarticletitle{ReMEmbR: Building and Reasoning Over Long-Horizon Spatio-Temporal Memory for Robot Navigation}.
\newblock \bibinfo{journal}{\emph{arXiv preprint arXiv:2409.13682}} (\bibinfo{year}{2024}).
\newblock


\bibitem[Bhatlawande et~al\mbox{.}(2012)]%
        {headmount}
\bibfield{author}{\bibinfo{person}{Shripad~S. Bhatlawande}, \bibinfo{person}{Jayant Mukhopadhyay}, {and} \bibinfo{person}{Manjunatha Mahadevappa}.} \bibinfo{year}{2012}\natexlab{}.
\newblock \showarticletitle{Ultrasonic spectacles and waist-belt for visually impaired and blind person}. In \bibinfo{booktitle}{\emph{2012 National Conference on Communications (NCC)}}. \bibinfo{pages}{1--4}.
\newblock
\urldef\tempurl%
\url{https://doi.org/10.1109/NCC.2012.6176765}
\showDOI{\tempurl}


\bibitem[Biswas(2014)]%
        {biswas2014vector}
\bibfield{author}{\bibinfo{person}{Joydeep Biswas}.} \bibinfo{year}{2014}\natexlab{}.
\newblock \emph{\bibinfo{title}{Vector Map-Based, Non-Markov Localization for Long-Term Deployment of Autonomous Mobile Robots}}.
\newblock PhD Thesis. \bibinfo{school}{Carnegie Mellon University}.
\newblock
\urldef\tempurl%
\url{https://joydeepb.com/Publications/joydeepb_thesis.pdf}
\showURL{%
\tempurl}


\bibitem[Biswas and Veloso(2013)]%
        {biswas2013localization}
\bibfield{author}{\bibinfo{person}{Joydeep Biswas} {and} \bibinfo{person}{Manuela~M Veloso}.} \bibinfo{year}{2013}\natexlab{}.
\newblock \showarticletitle{Localization and navigation of the cobots over long-term deployments}.
\newblock \bibinfo{journal}{\emph{The International Journal of Robotics Research}} \bibinfo{volume}{32}, \bibinfo{number}{14} (\bibinfo{year}{2013}), \bibinfo{pages}{1679--1694}.
\newblock


\bibitem[Cai et~al\mbox{.}(2024)]%
        {cai2024navigating}
\bibfield{author}{\bibinfo{person}{Shaojun Cai}, \bibinfo{person}{Ashwin Ram}, \bibinfo{person}{Zhengtai Gou}, \bibinfo{person}{Mohd Alqama~Wasim Shaikh}, \bibinfo{person}{Yu-An Chen}, \bibinfo{person}{Yingjia Wan}, \bibinfo{person}{Kotaro Hara}, \bibinfo{person}{Shengdong Zhao}, {and} \bibinfo{person}{David Hsu}.} \bibinfo{year}{2024}\natexlab{}.
\newblock \showarticletitle{Navigating Real-World Challenges: A Quadruped Robot Guiding System for Visually Impaired People in Diverse Environments}. In \bibinfo{booktitle}{\emph{Proceedings of the CHI Conference on Human Factors in Computing Systems}}. \bibinfo{pages}{1--18}.
\newblock


\bibitem[Chen et~al\mbox{.}(2022a)]%
        {chen2022can}
\bibfield{author}{\bibinfo{person}{Qihe Chen}, \bibinfo{person}{Luyao Wang}, \bibinfo{person}{Yan Zhang}, \bibinfo{person}{ZiangL Li}, \bibinfo{person}{Tingmin Yan}, \bibinfo{person}{Fan Wang}, \bibinfo{person}{Guyue Zhou}, {and} \bibinfo{person}{Jiangtao Gong}.} \bibinfo{year}{2022}\natexlab{a}.
\newblock \showarticletitle{Can Quadruped Navigation Robots be Used as Guide Dogs?}
\newblock \bibinfo{journal}{\emph{arXiv preprint arXiv:2210.08727}} (\bibinfo{year}{2022}).
\newblock


\bibitem[Chen et~al\mbox{.}(2022b)]%
        {chen2022quadruped}
\bibfield{author}{\bibinfo{person}{Yanbo Chen}, \bibinfo{person}{Zhengzhe Xu}, \bibinfo{person}{Zhuozhu Jian}, \bibinfo{person}{Gengpan Tang}, \bibinfo{person}{Yunong Yangli}, \bibinfo{person}{Anxing Xiao}, \bibinfo{person}{Xueqian Wang}, {and} \bibinfo{person}{Bin Liang}.} \bibinfo{year}{2022}\natexlab{b}.
\newblock \showarticletitle{Quadruped Guidance Robot for the Visually Impaired: A Comfort-Based Approach}.
\newblock \bibinfo{journal}{\emph{arXiv preprint arXiv:2203.03927}} (\bibinfo{year}{2022}).
\newblock


\bibitem[Cheng et~al\mbox{.}(2023)]%
        {cheng2023parkour}
\bibfield{author}{\bibinfo{person}{Xuxin Cheng}, \bibinfo{person}{Kexin Shi}, \bibinfo{person}{Ananye Agarwal}, {and} \bibinfo{person}{Deepak Pathak}.} \bibinfo{year}{2023}\natexlab{}.
\newblock \showarticletitle{Extreme Parkour with Legged Robots}.
\newblock \bibinfo{journal}{\emph{arXiv preprint arXiv:2309.14341}} (\bibinfo{year}{2023}).
\newblock


\bibitem[Chuang et~al\mbox{.}(2018)]%
        {chuang2018deep}
\bibfield{author}{\bibinfo{person}{Tzu-Kuan Chuang}, \bibinfo{person}{Ni-Ching Lin}, \bibinfo{person}{Jih-Shi Chen}, \bibinfo{person}{Chen-Hao Hung}, \bibinfo{person}{Yi-Wei Huang}, \bibinfo{person}{Chunchih Teng}, \bibinfo{person}{Haikun Huang}, \bibinfo{person}{Lap-Fai Yu}, \bibinfo{person}{Laura Giarr{\'e}}, {and} \bibinfo{person}{Hsueh-Cheng Wang}.} \bibinfo{year}{2018}\natexlab{}.
\newblock \showarticletitle{Deep trail-following robotic guide dog in pedestrian environments for people who are blind and visually impaired-learning from virtual and real worlds}. In \bibinfo{booktitle}{\emph{2018 IEEE International Conference on Robotics and Automation (ICRA)}}. IEEE, \bibinfo{pages}{5849--5855}.
\newblock


\bibitem[Clark-Carter et~al\mbox{.}(1986)]%
        {clark1986efficiency}
\bibfield{author}{\bibinfo{person}{DD Clark-Carter}, \bibinfo{person}{AD Heyes}, {and} \bibinfo{person}{CI Howarth}.} \bibinfo{year}{1986}\natexlab{}.
\newblock \showarticletitle{The efficiency and walking speed of visually impaired people}.
\newblock \bibinfo{journal}{\emph{Ergonomics}} \bibinfo{volume}{29}, \bibinfo{number}{6} (\bibinfo{year}{1986}), \bibinfo{pages}{779--789}.
\newblock


\bibitem[Dakopoulos and Bourbakis(2009)]%
        {dakopoulos2009wearable}
\bibfield{author}{\bibinfo{person}{Dimitrios Dakopoulos} {and} \bibinfo{person}{Nikolaos~G Bourbakis}.} \bibinfo{year}{2009}\natexlab{}.
\newblock \showarticletitle{Wearable obstacle avoidance electronic travel aids for blind: a survey}.
\newblock \bibinfo{journal}{\emph{IEEE Transactions on Systems, Man, and Cybernetics, Part C (Applications and Reviews)}} \bibinfo{volume}{40}, \bibinfo{number}{1} (\bibinfo{year}{2009}), \bibinfo{pages}{25--35}.
\newblock


\bibitem[DeepRobotics({[n.\,d.]})]%
        {deeprobotics_lite3}
\bibfield{author}{\bibinfo{person}{DeepRobotics}.} \bibinfo{year}{[n.\,d.]}\natexlab{}.
\newblock \bibinfo{title}{The Lite 3 Series Advanced Bionic Robotic Dog}.
\newblock \bibinfo{howpublished}{\url{https://www.deeprobotics.cn/en/index/product1.html}}.
\newblock
\newblock
\shownote{Accessed: 2023-12-01}.


\bibitem[Due(2023)]%
        {due2023walk}
\bibfield{author}{\bibinfo{person}{Brian~L Due}.} \bibinfo{year}{2023}\natexlab{}.
\newblock \showarticletitle{A walk in the park with Robodog: Navigating around pedestrians using a spot robot as a “guide dog”}.
\newblock \bibinfo{journal}{\emph{Space and Culture}} (\bibinfo{year}{2023}), \bibinfo{pages}{12063312231159215}.
\newblock


\bibitem[Faria et~al\mbox{.}(2010)]%
        {faria2010electronic}
\bibfield{author}{\bibinfo{person}{Jos{\'e} Faria}, \bibinfo{person}{S{\'e}rgio Lopes}, \bibinfo{person}{Hugo Fernandes}, \bibinfo{person}{Paulo Martins}, {and} \bibinfo{person}{Jo{\~a}o Barroso}.} \bibinfo{year}{2010}\natexlab{}.
\newblock \showarticletitle{Electronic white cane for blind people navigation assistance}. In \bibinfo{booktitle}{\emph{2010 World Automation Congress}}. IEEE, \bibinfo{pages}{1--7}.
\newblock


\bibitem[for Disease~Control and Prevention(2020)]%
        {cdc_fast_facts}
\bibfield{author}{\bibinfo{person}{Centers for Disease~Control} {and} \bibinfo{person}{Prevention}.} \bibinfo{year}{2020}\natexlab{}.
\newblock \bibinfo{title}{Fast Facts of Common Eye Disorders}.
\newblock \bibinfo{howpublished}{\url{https://www.cdc.gov/visionhealth/basics/ced/fastfacts.htm}}.
\newblock


\bibitem[for~the Blind({[n.\,d.]})]%
        {howmany-guidedog}
\bibfield{author}{\bibinfo{person}{Guiding~Eyes for~the Blind}.} \bibinfo{year}{[n.\,d.]}\natexlab{}.
\newblock \bibinfo{title}{{FAQs}}.
\newblock \bibinfo{howpublished}{\url{https://www.guidingeyes.org/about/faqs/}}.
\newblock


\bibitem[Forbes({[n.\,d.]})]%
        {forbes_guidedog}
\bibfield{author}{\bibinfo{person}{Forbes}.} \bibinfo{year}{[n.\,d.]}\natexlab{}.
\newblock \bibinfo{title}{Here's How To Get A Guide Dog}.
\newblock \bibinfo{howpublished}{\url{https://www.forbes.com/sites/peterslatin/2019/07/03/heres-how-to-get-a-guide-dog/?sh=2a6415e4584a}}.
\newblock
\newblock
\shownote{Accessed: 2023-3-17}.


\bibitem[Freedy et~al\mbox{.}(2007)]%
        {freedy2007measurement}
\bibfield{author}{\bibinfo{person}{Amos Freedy}, \bibinfo{person}{Ewart DeVisser}, \bibinfo{person}{Gershon Weltman}, {and} \bibinfo{person}{Nicole Coeyman}.} \bibinfo{year}{2007}\natexlab{}.
\newblock \showarticletitle{Measurement of trust in human-robot collaboration}. In \bibinfo{booktitle}{\emph{2007 International symposium on collaborative technologies and systems}}. Ieee, \bibinfo{pages}{106--114}.
\newblock


\bibitem[Galatas et~al\mbox{.}(2011)]%
        {galatas2011eyedog}
\bibfield{author}{\bibinfo{person}{Georgios Galatas}, \bibinfo{person}{Christopher McMurrough}, \bibinfo{person}{Gian~Luca Mariottini}, {and} \bibinfo{person}{Fillia Makedon}.} \bibinfo{year}{2011}\natexlab{}.
\newblock \showarticletitle{eyeDog: an assistive-guide robot for the visually impaired}. In \bibinfo{booktitle}{\emph{Proceedings of the 4th international conference on pervasive technologies related to assistive environments}}. \bibinfo{pages}{1--8}.
\newblock


\bibitem[Geiger et~al\mbox{.}(2012)]%
        {geiger2012we}
\bibfield{author}{\bibinfo{person}{Andreas Geiger}, \bibinfo{person}{Philip Lenz}, {and} \bibinfo{person}{Raquel Urtasun}.} \bibinfo{year}{2012}\natexlab{}.
\newblock \showarticletitle{Are we ready for autonomous driving? the kitti vision benchmark suite}. In \bibinfo{booktitle}{\emph{2012 IEEE conference on computer vision and pattern recognition}}. IEEE, \bibinfo{pages}{3354--3361}.
\newblock


\bibitem[Giudice and Long(2010)]%
        {giudice2010establishing}
\bibfield{author}{\bibinfo{person}{NA Giudice} {and} \bibinfo{person}{RG Long}.} \bibinfo{year}{2010}\natexlab{}.
\newblock \showarticletitle{Establishing and maintaining orientation: tools, techniques, and technologies}.
\newblock \bibinfo{journal}{\emph{Foundations of Orientation and Mobility, 4th ed.; APH Press: Louisville, KY, USA}}  \bibinfo{volume}{1} (\bibinfo{year}{2010}), \bibinfo{pages}{45--62}.
\newblock


\bibitem[Gothard et~al\mbox{.}(1996)]%
        {gothard1996dynamics}
\bibfield{author}{\bibinfo{person}{Katalin~M Gothard}, \bibinfo{person}{William~E Skaggs}, {and} \bibinfo{person}{Bruce~L McNaughton}.} \bibinfo{year}{1996}\natexlab{}.
\newblock \showarticletitle{Dynamics of mismatch correction in the hippocampal ensemble code for space: interaction between path integration and environmental cues}.
\newblock \bibinfo{journal}{\emph{Journal of Neuroscience}} \bibinfo{volume}{16}, \bibinfo{number}{24} (\bibinfo{year}{1996}), \bibinfo{pages}{8027--8040}.
\newblock


\bibitem[Guerreiro et~al\mbox{.}(2019)]%
        {guerreiro2019cabot}
\bibfield{author}{\bibinfo{person}{Jo{\~a}o Guerreiro}, \bibinfo{person}{Daisuke Sato}, \bibinfo{person}{Saki Asakawa}, \bibinfo{person}{Huixu Dong}, \bibinfo{person}{Kris~M Kitani}, {and} \bibinfo{person}{Chieko Asakawa}.} \bibinfo{year}{2019}\natexlab{}.
\newblock \showarticletitle{Cabot: Designing and evaluating an autonomous navigation robot for blind people}. In \bibinfo{booktitle}{\emph{The 21st International ACM SIGACCESS conference on computers and accessibility}}. \bibinfo{pages}{68--82}.
\newblock


\bibitem[Hamed et~al\mbox{.}(2019)]%
        {hamed2019hierarchical}
\bibfield{author}{\bibinfo{person}{Kaveh~Akbari Hamed}, \bibinfo{person}{Vinay~R Kamidi}, \bibinfo{person}{Wen-Loong Ma}, \bibinfo{person}{Alexander Leonessa}, {and} \bibinfo{person}{Aaron~D Ames}.} \bibinfo{year}{2019}\natexlab{}.
\newblock \showarticletitle{Hierarchical and safe motion control for cooperative locomotion of robotic guide dogs and humans: A hybrid systems approach}.
\newblock \bibinfo{journal}{\emph{IEEE Robotics and Automation Letters}} \bibinfo{volume}{5}, \bibinfo{number}{1} (\bibinfo{year}{2019}), \bibinfo{pages}{56--63}.
\newblock


\bibitem[Hersh and Johnson(2010)]%
        {hersh2010robotic}
\bibfield{author}{\bibinfo{person}{Marion~A Hersh} {and} \bibinfo{person}{Michael~A Johnson}.} \bibinfo{year}{2010}\natexlab{}.
\newblock \showarticletitle{A robotic guide for blind people. Part 1. A multi-national survey of the attitudes, requirements and preferences of potential end-users}.
\newblock \bibinfo{journal}{\emph{Applied Bionics and Biomechanics}} \bibinfo{volume}{7}, \bibinfo{number}{4} (\bibinfo{year}{2010}), \bibinfo{pages}{277--288}.
\newblock


\bibitem[Hirose et~al\mbox{.}(2018)]%
        {GOnet}
\bibfield{author}{\bibinfo{person}{Noriaki Hirose}, \bibinfo{person}{Amir Sadeghian}, \bibinfo{person}{Marynel V{\'{a}}zquez}, \bibinfo{person}{Patrick Goebel}, {and} \bibinfo{person}{Silvio Savarese}.} \bibinfo{year}{2018}\natexlab{}.
\newblock \showarticletitle{GONet: {A} Semi-Supervised Deep Learning Approach For Traversability Estimation}.
\newblock \bibinfo{journal}{\emph{CoRR}}  \bibinfo{volume}{abs/1803.03254} (\bibinfo{year}{2018}).
\newblock
\showeprint[arXiv]{1803.03254}
\urldef\tempurl%
\url{http://arxiv.org/abs/1803.03254}
\showURL{%
\tempurl}


\bibitem[Hoeller et~al\mbox{.}(2023)]%
        {hoeller2023anymal}
\bibfield{author}{\bibinfo{person}{David Hoeller}, \bibinfo{person}{Nikita Rudin}, \bibinfo{person}{Dhionis Sako}, {and} \bibinfo{person}{Marco Hutter}.} \bibinfo{year}{2023}\natexlab{}.
\newblock \bibinfo{title}{ANYmal Parkour: Learning Agile Navigation for Quadrupedal Robots}.
\newblock
\newblock
\showeprint[arxiv]{2306.14874}~[cs.RO]


\bibitem[Hwang et~al\mbox{.}(2024a)]%
        {hwang2024towards}
\bibfield{author}{\bibinfo{person}{Hochul Hwang}, \bibinfo{person}{Hee-Tae Jung}, \bibinfo{person}{Nicholas~A Giudice}, \bibinfo{person}{Joydeep Biswas}, \bibinfo{person}{Sunghoon~Ivan Lee}, {and} \bibinfo{person}{Donghyun Kim}.} \bibinfo{year}{2024}\natexlab{a}.
\newblock \showarticletitle{Towards Robotic Companions: Understanding Handler-Guide Dog Interactions for Informed Guide Dog Robot Design}. In \bibinfo{booktitle}{\emph{Proceedings of the CHI Conference on Human Factors in Computing Systems}}. \bibinfo{pages}{1--20}.
\newblock


\bibitem[Hwang et~al\mbox{.}(2024b)]%
        {hwang2024safe}
\bibfield{author}{\bibinfo{person}{Hochul Hwang}, \bibinfo{person}{Sunjae Kwon}, \bibinfo{person}{Yekyung Kim}, {and} \bibinfo{person}{Donghyun Kim}.} \bibinfo{year}{2024}\natexlab{b}.
\newblock \showarticletitle{Is it safe to cross? Interpretable Risk Assessment with GPT-4V for Safety-Aware Street Crossing}.
\newblock \bibinfo{journal}{\emph{arXiv preprint arXiv:2402.06794}} (\bibinfo{year}{2024}).
\newblock


\bibitem[Hwang et~al\mbox{.}(2023)]%
        {hwang2023system}
\bibfield{author}{\bibinfo{person}{Hochul Hwang}, \bibinfo{person}{Tim Xia}, \bibinfo{person}{Ibrahima Keita}, \bibinfo{person}{Ken Suzuki}, \bibinfo{person}{Joydeep Biswas}, \bibinfo{person}{Sunghoon~I Lee}, {and} \bibinfo{person}{Donghyun Kim}.} \bibinfo{year}{2023}\natexlab{}.
\newblock \showarticletitle{System configuration and navigation of a guide dog robot: Toward animal guide dog-level guiding work}. In \bibinfo{booktitle}{\emph{2023 IEEE International Conference on Robotics and Automation (ICRA)}}. IEEE, \bibinfo{pages}{9778--9784}.
\newblock


\bibitem[Institute({[n.\,d.]})]%
        {nih_double}
\bibfield{author}{\bibinfo{person}{NIH National~Eye Institute}.} \bibinfo{year}{[n.\,d.]}\natexlab{}.
\newblock \bibinfo{title}{Visual impairment, blindness cases in U.S. expected to double by 2050}.
\newblock \bibinfo{howpublished}{\url{https://www.nei.nih.gov/about/news-and-events/news/visual-impairment-blindness-cases-us-expected-double-2050}}.
\newblock
\newblock
\shownote{Accessed: 2023-1-17}.


\bibitem[{International Guide Dog Federation}({[n.\,d.]})]%
        {History_Guide_Dogs}
\bibfield{author}{\bibinfo{person}{{International Guide Dog Federation}}.} \bibinfo{year}{[n.\,d.]}\natexlab{}.
\newblock \bibinfo{title}{{History of Guide Dogs}}.
\newblock \bibinfo{howpublished}{\url{https://www.igdf.org.uk/guide-dogs/history-of-guide-dogs/}}.
\newblock
\newblock
\shownote{Accessed: 2022-11-30}.


\bibitem[Katzschmann et~al\mbox{.}(2018)]%
        {katzschmann2018safe}
\bibfield{author}{\bibinfo{person}{Robert~K Katzschmann}, \bibinfo{person}{Brandon Araki}, {and} \bibinfo{person}{Daniela Rus}.} \bibinfo{year}{2018}\natexlab{}.
\newblock \showarticletitle{Safe local navigation for visually impaired users with a time-of-flight and haptic feedback device}.
\newblock \bibinfo{journal}{\emph{IEEE Transactions on Neural Systems and Rehabilitation Engineering}} \bibinfo{volume}{26}, \bibinfo{number}{3} (\bibinfo{year}{2018}), \bibinfo{pages}{583--593}.
\newblock


\bibitem[Kayukawa et~al\mbox{.}(2019)]%
        {kayukawa2019bbeep}
\bibfield{author}{\bibinfo{person}{Seita Kayukawa}, \bibinfo{person}{Keita Higuchi}, \bibinfo{person}{Jo{\~a}o Guerreiro}, \bibinfo{person}{Shigeo Morishima}, \bibinfo{person}{Yoichi Sato}, \bibinfo{person}{Kris Kitani}, {and} \bibinfo{person}{Chieko Asakawa}.} \bibinfo{year}{2019}\natexlab{}.
\newblock \showarticletitle{BBeep: A sonic collision avoidance system for blind travellers and nearby pedestrians}. In \bibinfo{booktitle}{\emph{Proceedings of the 2019 CHI Conference on Human Factors in Computing Systems}}. \bibinfo{pages}{1--12}.
\newblock


\bibitem[Kim et~al\mbox{.}(2023)]%
        {kim2023train}
\bibfield{author}{\bibinfo{person}{J~Taery Kim}, \bibinfo{person}{Wenhao Yu}, \bibinfo{person}{Jie Tan}, \bibinfo{person}{Greg Turk}, {and} \bibinfo{person}{Sehoon Ha}.} \bibinfo{year}{2023}\natexlab{}.
\newblock \showarticletitle{How to train your guide dog: Wayfinding and safe navigation with human-robot modeling}. In \bibinfo{booktitle}{\emph{Companion of the 2023 ACM/IEEE International Conference on Human-Robot Interaction}}. \bibinfo{pages}{221--225}.
\newblock


\bibitem[Kulyukin et~al\mbox{.}(2004)]%
        {kulyukin2004robotic}
\bibfield{author}{\bibinfo{person}{Vladimir Kulyukin}, \bibinfo{person}{Chaitanya Gharpure}, \bibinfo{person}{Pradnya Sute}, \bibinfo{person}{Nathan De~Graw}, \bibinfo{person}{John Nicholson}, {and} \bibinfo{person}{S Pavithran}.} \bibinfo{year}{2004}\natexlab{}.
\newblock \showarticletitle{A robotic wayfinding system for the visually impaired}. In \bibinfo{booktitle}{\emph{Proceedings of the National Conference on Artificial Intelligence}}. Menlo Park, CA; Cambridge, MA; London; AAAI Press; MIT Press; 1999, \bibinfo{pages}{864--869}.
\newblock


\bibitem[L\"a\"ane(2020)]%
        {starship}
\bibfield{author}{\bibinfo{person}{Joan L\"a\"ane}.} \bibinfo{year}{2020}\natexlab{}.
\newblock \bibinfo{title}{How Starship Delivery Robots know where they are going}.
\newblock \bibinfo{howpublished}{\url{https://medium.com/starshiptechnologies/how-starship-delivery-robots-know-where-they-are-going-c97d385a1015}}.
\newblock


\bibitem[Lee et~al\mbox{.}(2020)]%
        {lee2020learning}
\bibfield{author}{\bibinfo{person}{Joonho Lee}, \bibinfo{person}{Jemin Hwangbo}, \bibinfo{person}{Lorenz Wellhausen}, \bibinfo{person}{Vladlen Koltun}, {and} \bibinfo{person}{Marco Hutter}.} \bibinfo{year}{2020}\natexlab{}.
\newblock \showarticletitle{Learning quadrupedal locomotion over challenging terrain}.
\newblock \bibinfo{journal}{\emph{Science robotics}} \bibinfo{volume}{5}, \bibinfo{number}{47} (\bibinfo{year}{2020}), \bibinfo{pages}{eabc5986}.
\newblock


\bibitem[Li et~al\mbox{.}(2016)]%
        {li2016isana}
\bibfield{author}{\bibinfo{person}{Bing Li}, \bibinfo{person}{J~Pablo Munoz}, \bibinfo{person}{Xuejian Rong}, \bibinfo{person}{Jizhong Xiao}, \bibinfo{person}{Yingli Tian}, {and} \bibinfo{person}{Aries Arditi}.} \bibinfo{year}{2016}\natexlab{}.
\newblock \showarticletitle{ISANA: wearable context-aware indoor assistive navigation with obstacle avoidance for the blind}. In \bibinfo{booktitle}{\emph{Computer Vision--ECCV 2016 Workshops: Amsterdam, The Netherlands, October 8-10 and 15-16, 2016, Proceedings, Part II 14}}. Springer, \bibinfo{pages}{448--462}.
\newblock


\bibitem[Lloyd et~al\mbox{.}(2008)]%
        {lloyd2008guide}
\bibfield{author}{\bibinfo{person}{Janice~KF Lloyd}, \bibinfo{person}{Steven La~Grow}, \bibinfo{person}{Kevin~J Stafford}, {and} \bibinfo{person}{R~Claire Budge}.} \bibinfo{year}{2008}\natexlab{}.
\newblock \showarticletitle{The guide dog as a mobility aid part 2: Perceived changes to travel habits}.
\newblock \bibinfo{journal}{\emph{Vision Rehabilitation International}} \bibinfo{volume}{1}, \bibinfo{number}{1} (\bibinfo{year}{2008}), \bibinfo{pages}{34--45}.
\newblock


\bibitem[Macenski et~al\mbox{.}(2020)]%
        {macenski2020marathon2}
\bibfield{author}{\bibinfo{person}{Steven Macenski}, \bibinfo{person}{Francisco Martin}, \bibinfo{person}{Ruffin White}, {and} \bibinfo{person}{Jonatan Ginés~Clavero}.} \bibinfo{year}{2020}\natexlab{}.
\newblock \showarticletitle{The Marathon 2: A Navigation System}. In \bibinfo{booktitle}{\emph{2020 IEEE/RSJ International Conference on Intelligent Robots and Systems (IROS)}}.
\newblock


\bibitem[Macenski et~al\mbox{.}(2023)]%
        {macenski2023survey}
\bibfield{author}{\bibinfo{person}{S. Macenski}, \bibinfo{person}{T. Moore}, \bibinfo{person}{DV Lu}, \bibinfo{person}{A. Merzlyakov}, {and} \bibinfo{person}{M. Ferguson}.} \bibinfo{year}{2023}\natexlab{}.
\newblock \showarticletitle{{From the desks of ROS maintainers: A survey of modern \& capable mobile robotics algorithms in the robot operating system 2}}.
\newblock \bibinfo{journal}{\emph{Robotics and Autonomous Systems}} (\bibinfo{year}{2023}).
\newblock


\bibitem[Megalingam et~al\mbox{.}(2019)]%
        {megalingam2019autonomous}
\bibfield{author}{\bibinfo{person}{Rajesh~Kannan Megalingam}, \bibinfo{person}{Souraj Vishnu}, \bibinfo{person}{Vishnu Sasikumar}, {and} \bibinfo{person}{Sajikumar Sreekumar}.} \bibinfo{year}{2019}\natexlab{}.
\newblock \showarticletitle{Autonomous path guiding robot for visually impaired people}. In \bibinfo{booktitle}{\emph{Cognitive Informatics and Soft Computing: Proceeding of CISC 2017}}. Springer, \bibinfo{pages}{257--266}.
\newblock


\bibitem[Merzlyakov and Macenski(2021)]%
        {vslamComparison2021}
\bibfield{author}{\bibinfo{person}{Alexey Merzlyakov} {and} \bibinfo{person}{Steven Macenski}.} \bibinfo{year}{2021}\natexlab{}.
\newblock \showarticletitle{A Comparison of Modern General-Purpose Visual SLAM Approaches}. In \bibinfo{booktitle}{\emph{2021 IEEE/RSJ International Conference on Intelligent Robots and Systems (IROS)}}.
\newblock


\bibitem[Miki et~al\mbox{.}(2022)]%
        {miki2022learning}
\bibfield{author}{\bibinfo{person}{Takahiro Miki}, \bibinfo{person}{Joonho Lee}, \bibinfo{person}{Jemin Hwangbo}, \bibinfo{person}{Lorenz Wellhausen}, \bibinfo{person}{Vladlen Koltun}, {and} \bibinfo{person}{Marco Hutter}.} \bibinfo{year}{2022}\natexlab{}.
\newblock \showarticletitle{Learning robust perceptive locomotion for quadrupedal robots in the wild}.
\newblock \bibinfo{journal}{\emph{Science Robotics}} \bibinfo{volume}{7}, \bibinfo{number}{62} (\bibinfo{year}{2022}), \bibinfo{pages}{eabk2822}.
\newblock


\bibitem[Miner(2001)]%
        {miner2001experience}
\bibfield{author}{\bibinfo{person}{Rachel Joy-Taub Miner}.} \bibinfo{year}{2001}\natexlab{}.
\newblock \showarticletitle{The experience of living with and using a dog guide}.
\newblock \bibinfo{journal}{\emph{RE: view}} \bibinfo{volume}{32}, \bibinfo{number}{4} (\bibinfo{year}{2001}), \bibinfo{pages}{183}.
\newblock


\bibitem[my~eye({[n.\,d.]})]%
        {guide-dog-interview}
\bibfield{author}{\bibinfo{person}{Be my eye}.} \bibinfo{year}{[n.\,d.]}\natexlab{}.
\newblock \bibinfo{title}{White canes and guide dogs: what's actually the difference?}
\newblock \bibinfo{howpublished}{\url{https://www.bemyeyes.com/podcasts/white-canes-and-guide-dogs-whats-actually-the-difference}}.
\newblock
\newblock
\shownote{Accessed: 2023-1-17}.


\bibitem[Nanavati et~al\mbox{.}(2018)]%
        {nanavati2018coupled}
\bibfield{author}{\bibinfo{person}{Amal Nanavati}, \bibinfo{person}{Xiang~Zhi Tan}, {and} \bibinfo{person}{Aaron Steinfeld}.} \bibinfo{year}{2018}\natexlab{}.
\newblock \showarticletitle{Coupled indoor navigation for people who are blind}. In \bibinfo{booktitle}{\emph{Companion of the 2018 ACM/IEEE International Conference on Human-Robot Interaction}}. \bibinfo{pages}{201--202}.
\newblock


\bibitem[{NSK Ltd}({[n.\,d.]})]%
        {NSK}
\bibfield{author}{\bibinfo{person}{{NSK Ltd}}.} \bibinfo{year}{[n.\,d.]}\natexlab{}.
\newblock \bibinfo{title}{NSK Develops a Guide-Dog Style Robot}.
\newblock \bibinfo{howpublished}{\url{https://www.nsk.com/company/news/2011/press111027b.html}}.
\newblock
\newblock
\shownote{Accessed: 2023-1-17}.


\bibitem[Oquab et~al\mbox{.}(2023)]%
        {oquab2023dinov2}
\bibfield{author}{\bibinfo{person}{Maxime Oquab}, \bibinfo{person}{Timoth{\'e}e Darcet}, \bibinfo{person}{Th{\'e}o Moutakanni}, \bibinfo{person}{Huy Vo}, \bibinfo{person}{Marc Szafraniec}, \bibinfo{person}{Vasil Khalidov}, \bibinfo{person}{Pierre Fernandez}, \bibinfo{person}{Daniel Haziza}, \bibinfo{person}{Francisco Massa}, \bibinfo{person}{Alaaeldin El-Nouby}, {et~al\mbox{.}}} \bibinfo{year}{2023}\natexlab{}.
\newblock \showarticletitle{Dinov2: Learning robust visual features without supervision}.
\newblock \bibinfo{journal}{\emph{arXiv preprint arXiv:2304.07193}} (\bibinfo{year}{2023}).
\newblock


\bibitem[Prada and Forero(2022)]%
        {belt}
\bibfield{author}{\bibinfo{person}{Erick Javier~Argüello Prada} {and} \bibinfo{person}{Lina María~Santacruz Forero}.} \bibinfo{year}{2022}\natexlab{}.
\newblock \showarticletitle{A belt-like assistive device for visually impaired people: Toward a more collaborative approach}.
\newblock \bibinfo{journal}{\emph{Cogent Engineering}} \bibinfo{volume}{9}, \bibinfo{number}{1} (\bibinfo{year}{2022}), \bibinfo{pages}{2048440}.
\newblock
\urldef\tempurl%
\url{https://doi.org/10.1080/23311916.2022.2048440}
\showDOI{\tempurl}


\bibitem[Rabiee and Biswas(2020)]%
        {rabiee2020ivslam}
\bibfield{author}{\bibinfo{person}{Sadegh Rabiee} {and} \bibinfo{person}{Joydeep Biswas}.} \bibinfo{year}{2020}\natexlab{}.
\newblock \bibinfo{title}{{IV-SLAM: Introspective Vision for Simultaneous Localization and Mapping}}.
\newblock \bibinfo{howpublished}{arXiv Preprint arXiv:2008.02760}.
\newblock


\bibitem[Refson et~al\mbox{.}(1999)]%
        {refson1999health}
\bibfield{author}{\bibinfo{person}{K Refson}, \bibinfo{person}{AJ Jackson}, \bibinfo{person}{AE Dusoir}, {and} \bibinfo{person}{DB Archer}.} \bibinfo{year}{1999}\natexlab{}.
\newblock \showarticletitle{The health and social status of guide dog owners and other visually impaired adults in Scotland}.
\newblock \bibinfo{journal}{\emph{Visual Impairment Research}} \bibinfo{volume}{1}, \bibinfo{number}{2} (\bibinfo{year}{1999}), \bibinfo{pages}{95--109}.
\newblock


\bibitem[Robotics(2023)]%
        {UnitreeGo2}
\bibfield{author}{\bibinfo{person}{Unitree Robotics}.} \bibinfo{year}{2023}\natexlab{}.
\newblock \bibinfo{title}{Robot Dog Go 2}.
\newblock \bibinfo{howpublished}{\url{https://www.unitree.com/en/go2/}}.
\newblock


\bibitem[Rodr{\'\i}guez et~al\mbox{.}(2012)]%
        {rodriguez2012assisting}
\bibfield{author}{\bibinfo{person}{Alberto Rodr{\'\i}guez}, \bibinfo{person}{J~Javier Yebes}, \bibinfo{person}{Pablo~F Alcantarilla}, \bibinfo{person}{Luis~M Bergasa}, \bibinfo{person}{Javier Almaz{\'a}n}, {and} \bibinfo{person}{Andr{\'e}s Cela}.} \bibinfo{year}{2012}\natexlab{}.
\newblock \showarticletitle{Assisting the visually impaired: obstacle detection and warning system by acoustic feedback}.
\newblock \bibinfo{journal}{\emph{Sensors}} \bibinfo{volume}{12}, \bibinfo{number}{12} (\bibinfo{year}{2012}), \bibinfo{pages}{17476--17496}.
\newblock


\bibitem[Rudin et~al\mbox{.}(2022)]%
        {rudin2022learning}
\bibfield{author}{\bibinfo{person}{Nikita Rudin}, \bibinfo{person}{David Hoeller}, \bibinfo{person}{Philipp Reist}, {and} \bibinfo{person}{Marco Hutter}.} \bibinfo{year}{2022}\natexlab{}.
\newblock \showarticletitle{Learning to walk in minutes using massively parallel deep reinforcement learning}. In \bibinfo{booktitle}{\emph{Conference on Robot Learning}}. PMLR, \bibinfo{pages}{91--100}.
\newblock


\bibitem[Saaid et~al\mbox{.}(2016)]%
        {saaid2016smart}
\bibfield{author}{\bibinfo{person}{Mohammad~Farid Saaid}, \bibinfo{person}{AM Mohammad}, {and} \bibinfo{person}{MSA~Megat Ali}.} \bibinfo{year}{2016}\natexlab{}.
\newblock \showarticletitle{Smart cane with range notification for blind people}. In \bibinfo{booktitle}{\emph{2016 IEEE International Conference on Automatic Control and Intelligent Systems (I2CACIS)}}. IEEE, \bibinfo{pages}{225--229}.
\newblock


\bibitem[Shah et~al\mbox{.}(2023)]%
        {shah2023lm}
\bibfield{author}{\bibinfo{person}{Dhruv Shah}, \bibinfo{person}{B{\l}a{\.z}ej Osi{\'n}ski}, \bibinfo{person}{Sergey Levine}, {et~al\mbox{.}}} \bibinfo{year}{2023}\natexlab{}.
\newblock \showarticletitle{Lm-nav: Robotic navigation with large pre-trained models of language, vision, and action}. In \bibinfo{booktitle}{\emph{Conference on robot learning}}. PMLR, \bibinfo{pages}{492--504}.
\newblock


\bibitem[Slade et~al\mbox{.}(2021)]%
        {slade2021multimodal}
\bibfield{author}{\bibinfo{person}{Patrick Slade}, \bibinfo{person}{Arjun Tambe}, {and} \bibinfo{person}{Mykel~J Kochenderfer}.} \bibinfo{year}{2021}\natexlab{}.
\newblock \showarticletitle{Multimodal sensing and intuitive steering assistance improve navigation and mobility for people with impaired vision}.
\newblock \bibinfo{journal}{\emph{Science Robotics}} \bibinfo{volume}{6}, \bibinfo{number}{59} (\bibinfo{year}{2021}), \bibinfo{pages}{eabg6594}.
\newblock


\bibitem[Strumillo et~al\mbox{.}(2018)]%
        {strumillo2018different}
\bibfield{author}{\bibinfo{person}{Pawel Strumillo}, \bibinfo{person}{Michal Bujacz}, \bibinfo{person}{Przemyslaw Baranski}, \bibinfo{person}{Piotr Skulimowski}, \bibinfo{person}{Piotr Korbel}, \bibinfo{person}{Mateusz Owczarek}, \bibinfo{person}{Krzysztof Tomalczyk}, \bibinfo{person}{Alin Moldoveanu}, {and} \bibinfo{person}{Runar Unnthorsson}.} \bibinfo{year}{2018}\natexlab{}.
\newblock \showarticletitle{Different approaches to aiding blind persons in mobility and navigation in the “Naviton” and “Sound of Vision” projects}.
\newblock \bibinfo{journal}{\emph{Mobility of Visually Impaired People: Fundamentals and ICT Assistive Technologies}} (\bibinfo{year}{2018}), \bibinfo{pages}{435--468}.
\newblock


\bibitem[Tachi and Komoriya(1985)]%
        {Tachi}
\bibfield{author}{\bibinfo{person}{Susumu Tachi} {and} \bibinfo{person}{Kiyoshi Komoriya}.} \bibinfo{year}{1985}\natexlab{}.
\newblock \showarticletitle{Guide Dog Robot}.
\newblock \bibinfo{journal}{\emph{The Robotics Research 2 (The Second International Symposium 1984)}}  \bibinfo{volume}{2} (\bibinfo{year}{1985}), \bibinfo{pages}{333--349}.
\newblock


\bibitem[{Tachi Laboratory, The University of Tokyo}({[n.\,d.]})]%
        {meldog}
\bibfield{author}{\bibinfo{person}{{Tachi Laboratory, The University of Tokyo}}.} \bibinfo{year}{[n.\,d.]}\natexlab{}.
\newblock \bibinfo{title}{Guide Dog Robot (MELDOG)}.
\newblock \bibinfo{howpublished}{\url{https://tachilab.org/en/projects/meldog.html}}.
\newblock
\newblock
\shownote{Accessed: 2023-1-17}.


\bibitem[Takizawa et~al\mbox{.}(2012)]%
        {takizawa2012kinect}
\bibfield{author}{\bibinfo{person}{Hotaka Takizawa}, \bibinfo{person}{Shotaro Yamaguchi}, \bibinfo{person}{Mayumi Aoyagi}, \bibinfo{person}{Nobuo Ezaki}, {and} \bibinfo{person}{Shinji Mizuno}.} \bibinfo{year}{2012}\natexlab{}.
\newblock \showarticletitle{Kinect cane: An assistive system for the visually impaired based on three-dimensional object recognition}. In \bibinfo{booktitle}{\emph{2012 IEEE/SICE international symposium on system integration (SII)}}. IEEE, \bibinfo{pages}{740--745}.
\newblock


\bibitem[Takizawa et~al\mbox{.}(2015)]%
        {takizawa2015kinect}
\bibfield{author}{\bibinfo{person}{Hotaka Takizawa}, \bibinfo{person}{Shotaro Yamaguchi}, \bibinfo{person}{Mayumi Aoyagi}, \bibinfo{person}{Nobuo Ezaki}, {and} \bibinfo{person}{Shinji Mizuno}.} \bibinfo{year}{2015}\natexlab{}.
\newblock \showarticletitle{Kinect cane: An assistive system for the visually impaired based on the concept of object recognition aid}.
\newblock \bibinfo{journal}{\emph{Personal and Ubiquitous Computing}}  \bibinfo{volume}{19} (\bibinfo{year}{2015}), \bibinfo{pages}{955--965}.
\newblock


\bibitem[Tobita et~al\mbox{.}(2018)]%
        {tobita2018structure}
\bibfield{author}{\bibinfo{person}{Kazuteru Tobita}, \bibinfo{person}{Katsuyuki Sagayama}, \bibinfo{person}{Mayuko Mori}, {and} \bibinfo{person}{Ayako Tabuchi}.} \bibinfo{year}{2018}\natexlab{}.
\newblock \showarticletitle{Structure and examination of the guidance robot LIGHBOT for visually impaired and elderly people}.
\newblock \bibinfo{journal}{\emph{Journal of Robotics and Mechatronics}} \bibinfo{volume}{30}, \bibinfo{number}{1} (\bibinfo{year}{2018}), \bibinfo{pages}{86--92}.
\newblock


\bibitem[Tobita et~al\mbox{.}(2017)]%
        {tobita2017examination}
\bibfield{author}{\bibinfo{person}{Kazuteru Tobita}, \bibinfo{person}{Katsuyuki Sagayama}, {and} \bibinfo{person}{Hironori Ogawa}.} \bibinfo{year}{2017}\natexlab{}.
\newblock \showarticletitle{Examination of a guidance robot for visually impaired people}.
\newblock \bibinfo{journal}{\emph{Journal of Robotics and Mechatronics}} \bibinfo{volume}{29}, \bibinfo{number}{4} (\bibinfo{year}{2017}), \bibinfo{pages}{720--727}.
\newblock


\bibitem[Ulrich and Borenstein(2001)]%
        {ulrich2001guidecane}
\bibfield{author}{\bibinfo{person}{Iwan Ulrich} {and} \bibinfo{person}{Johann Borenstein}.} \bibinfo{year}{2001}\natexlab{}.
\newblock \showarticletitle{The GuideCane-applying mobile robot technologies to assist the visually impaired}.
\newblock \bibinfo{journal}{\emph{IEEE Transactions on Systems, Man, and Cybernetics-Part A: Systems and Humans}} \bibinfo{volume}{31}, \bibinfo{number}{2} (\bibinfo{year}{2001}), \bibinfo{pages}{131--136}.
\newblock


\bibitem[{Unitree Robotics}({[n.\,d.]})]%
        {unitree_go1}
\bibfield{author}{\bibinfo{person}{{Unitree Robotics}}.} \bibinfo{year}{[n.\,d.]}\natexlab{}.
\newblock \bibinfo{title}{{Go 1}}.
\newblock \bibinfo{howpublished}{https://shop.unitree.com/products/unitreeyushutechnologydog-artificial-intelligence-companion-bionic-companion-intelligent-robot-go1-quadruped-robot-dog}.
\newblock
\newblock
\shownote{Accessed: 2023-1-15}.


\bibitem[Wahab et~al\mbox{.}(2011)]%
        {smartcane}
\bibfield{author}{\bibinfo{person}{Mohd Helmy~Abd Wahab}, \bibinfo{person}{Amirul~A. Talib}, \bibinfo{person}{Herdawatie~Abdul Kadir}, \bibinfo{person}{Ayob Johari}, \bibinfo{person}{Ahmad Noraziah}, \bibinfo{person}{Roslina~Mohd Sidek}, {and} \bibinfo{person}{Ariffin~Abdul Mutalib}.} \bibinfo{year}{2011}\natexlab{}.
\newblock \showarticletitle{Smart Cane: Assistive Cane for Visually-impaired People}.
\newblock \bibinfo{journal}{\emph{CoRR}}  \bibinfo{volume}{abs/1110.5156} (\bibinfo{year}{2011}).
\newblock
\showeprint[arXiv]{1110.5156}
\urldef\tempurl%
\url{http://arxiv.org/abs/1110.5156}
\showURL{%
\tempurl}


\bibitem[Whitmarsh(2005)]%
        {whitmarsh2005benefits}
\bibfield{author}{\bibinfo{person}{Lorraine Whitmarsh}.} \bibinfo{year}{2005}\natexlab{}.
\newblock \showarticletitle{The benefits of guide dog ownership}.
\newblock \bibinfo{journal}{\emph{Visual impairment research}} \bibinfo{volume}{7}, \bibinfo{number}{1} (\bibinfo{year}{2005}), \bibinfo{pages}{27--42}.
\newblock


\bibitem[Wiener et~al\mbox{.}(2010)]%
        {wiener2010foundations}
\bibfield{author}{\bibinfo{person}{William~R Wiener}, \bibinfo{person}{Richard~L Welsh}, {and} \bibinfo{person}{Bruce~B Blasch}.} \bibinfo{year}{2010}\natexlab{}.
\newblock \bibinfo{booktitle}{\emph{Foundations of orientation and mobility}}. Vol.~\bibinfo{volume}{1}.
\newblock \bibinfo{publisher}{American Foundation for the Blind}.
\newblock


\bibitem[Xiao et~al\mbox{.}(2021)]%
        {xiao2021robotic}
\bibfield{author}{\bibinfo{person}{Anxing Xiao}, \bibinfo{person}{Wenzhe Tong}, \bibinfo{person}{Lizhi Yang}, \bibinfo{person}{Jun Zeng}, \bibinfo{person}{Zhongyu Li}, {and} \bibinfo{person}{Koushil Sreenath}.} \bibinfo{year}{2021}\natexlab{}.
\newblock \showarticletitle{Robotic guide dog: Leading a human with leash-guided hybrid physical interaction}. In \bibinfo{booktitle}{\emph{2021 IEEE International Conference on Robotics and Automation (ICRA)}}. IEEE, \bibinfo{pages}{11470--11476}.
\newblock


\bibitem[Xiaomi({[n.\,d.]})]%
        {xiaomi_cyberdog}
\bibfield{author}{\bibinfo{person}{Xiaomi}.} \bibinfo{year}{[n.\,d.]}\natexlab{}.
\newblock \bibinfo{title}{Xiaomi Launches CyberDog – An Open Source Quadruped Robot Companion}.
\newblock \bibinfo{howpublished}{\url{https://www.mi.com/global/discover/article?id=2069}}.
\newblock
\newblock
\shownote{Accessed: 2023-12-01}.


\bibitem[Ye et~al\mbox{.}(2016)]%
        {ye2016co}
\bibfield{author}{\bibinfo{person}{Cang Ye}, \bibinfo{person}{Soonhac Hong}, \bibinfo{person}{Xiangfei Qian}, {and} \bibinfo{person}{Wei Wu}.} \bibinfo{year}{2016}\natexlab{}.
\newblock \showarticletitle{Co-robotic cane: A new robotic navigation aid for the visually impaired}.
\newblock \bibinfo{journal}{\emph{IEEE Systems, Man, and Cybernetics Magazine}} \bibinfo{volume}{2}, \bibinfo{number}{2} (\bibinfo{year}{2016}), \bibinfo{pages}{33--42}.
\newblock


\bibitem[Zeng et~al\mbox{.}(2017)]%
        {zeng2017camera}
\bibfield{author}{\bibinfo{person}{Limin Zeng}, \bibinfo{person}{Markus Simros}, {and} \bibinfo{person}{Gerhard Weber}.} \bibinfo{year}{2017}\natexlab{}.
\newblock \showarticletitle{Camera-based mobile electronic travel aids support for cognitive mapping of unknown spaces}. In \bibinfo{booktitle}{\emph{Proceedings of the 19th international conference on human-computer interaction with mobile devices and services}}. \bibinfo{pages}{1--10}.
\newblock


\bibitem[Zhang and Kleeman(2009)]%
        {zhang2009robust}
\bibfield{author}{\bibinfo{person}{Alan~M Zhang} {and} \bibinfo{person}{Lindsay Kleeman}.} \bibinfo{year}{2009}\natexlab{}.
\newblock \showarticletitle{Robust appearance based visual route following for navigation in large-scale outdoor environments}.
\newblock \bibinfo{journal}{\emph{The International Journal of Robotics Research}} \bibinfo{volume}{28}, \bibinfo{number}{3} (\bibinfo{year}{2009}), \bibinfo{pages}{331--356}.
\newblock


\bibitem[Zhang et~al\mbox{.}(2023)]%
        {zhang2023follower}
\bibfield{author}{\bibinfo{person}{Yan Zhang}, \bibinfo{person}{Ziang Li}, \bibinfo{person}{Haole Guo}, \bibinfo{person}{Luyao Wang}, \bibinfo{person}{Qihe Chen}, \bibinfo{person}{Wenjie Jiang}, \bibinfo{person}{Mingming Fan}, \bibinfo{person}{Guyue Zhou}, {and} \bibinfo{person}{Jiangtao Gong}.} \bibinfo{year}{2023}\natexlab{}.
\newblock \showarticletitle{" I am the follower, also the boss": Exploring Different Levels of Autonomy and Machine Forms of Guiding Robots for the Visually Impaired}. In \bibinfo{booktitle}{\emph{Proceedings of the 2023 CHI Conference on Human Factors in Computing Systems}}. \bibinfo{pages}{1--22}.
\newblock


\bibitem[Zhu et~al\mbox{.}(2021)]%
        {zhu2021lifelong}
\bibfield{author}{\bibinfo{person}{Shifan Zhu}, \bibinfo{person}{Xinyu Zhang}, \bibinfo{person}{Shichun Guo}, \bibinfo{person}{Jun Li}, {and} \bibinfo{person}{Huaping Liu}.} \bibinfo{year}{2021}\natexlab{}.
\newblock \showarticletitle{Lifelong localization in semi-dynamic environment}. In \bibinfo{booktitle}{\emph{2021 IEEE International Conference on Robotics and Automation (ICRA)}}. IEEE, \bibinfo{pages}{14389--14395}.
\newblock


\bibitem[Zhuang et~al\mbox{.}(2023)]%
        {zhuang2023robot}
\bibfield{author}{\bibinfo{person}{Ziwen Zhuang}, \bibinfo{person}{Zipeng Fu}, \bibinfo{person}{Jianren Wang}, \bibinfo{person}{Christopher Atkeson}, \bibinfo{person}{Sören Schwertfeger}, \bibinfo{person}{Chelsea Finn}, {and} \bibinfo{person}{Hang Zhao}.} \bibinfo{year}{2023}\natexlab{}.
\newblock \showarticletitle{Robot Parkour Learning}. In \bibinfo{booktitle}{\emph{Conference on Robot Learning ({CoRL})}}.
\newblock


\end{thebibliography}

% \appendix




\end{document}